\documentclass[11pt]{article}

\usepackage[preprint]{acl}

\usepackage{times}
\usepackage{latexsym}

\usepackage[T1]{fontenc}

\usepackage[utf8]{inputenc}

\usepackage{microtype}

\usepackage{inconsolata}

\usepackage{booktabs}   
\usepackage{graphicx}

\usepackage{enumitem}

\usepackage{amsmath}
\usepackage{tikz}
\usepackage{xcolor}
\usepackage{enumitem}

\usepackage[most]{tcolorbox}
\newtcolorbox{hlbox}{colback=yellow!35, boxrule=0pt, arc=0pt, left=1pt, right=1pt, top=1pt, bottom=1pt, breakable, enhanced jigsaw}
  
\usetikzlibrary{arrows.meta,positioning}

\title{Attacking Only Half the Data: Two-Sided Evaluation of AI-Text Detector Robustness }

\author{Apu Chandraw Shill \\
  Verbex.ai  \\
  \texttt{apu3660@gmail.com} \\\And
  Rabindra Nath Nandi \\
  Independent Researcher \\
  \texttt{rabindro.rath@gmail.com} \\}

\begin{document}
\maketitle
\begin{abstract}
We investigate the robustness of widely used AI text detectors under attacks applied to both human and machine text. Prior evaluations typically attack only machine text, overlooking that human text is routinely modified through grammar correction, formatting, or invisible character insertion. We introduce a two-sided protocol that applies attacks to both classes. Across two corpora and nine detectors, reported AUROC gains vanish under two-sided evaluation, with attacked human text misclassified at rates up to $100\%$. While simple preprocessing reverses character-level attacks, word-level synonym substitution remains effective, meaning-preserving, and irreversible.
\end{abstract}

\section{Introduction}
Machine-generated text detectors are increasingly deployed in high-stakes contexts such as academic integrity checks, content moderation, and hiring. Yet their reliability has been questioned from the very beginning \citep{krishna2023paraphrasing}. A central problem is that detectors often fail when text is altered, even slightly, by paraphrasing, grammar correction, or invisible character insertion. These failures matter because text is rarely submitted in a perfectly ``clean'' form.

Prior work has largely focused on attacks against machine-generated text, comparing it to untouched human writing. For example, paraphrasing has been shown to collapse detector performance \citep{krishna2023paraphrasing}, and subsequent studies argued that this weakness is fundamental \citep{sadasivan2023canai}. Benchmarks such as RAID \citep{dugan2024raid} have made paraphrase the standard stress test. More recent work has begun to explore the human side: polishing human text with an LLM can cause detectors to misclassify it \citep{saha2025polished}, while DetectRL \citep{wu2024detectrl} attacks human text but evaluates against clean machine text. However, these designs remain one-sided, either attacking only machine or only human text, and often re-tuning thresholds so that false positives never surface. As a result, they miss the full picture.

In this work, we address the missing half. We introduce a two-sided evaluation protocol that applies attacks to both human and machine text and scores them against each other. This design reveals failures that are invisible under one-sided evaluation. For example, under zero-width insertion, detectors flag up to $98\%$ of human documents while appearing to improve AUROC when only machine text is considered. The apparent ``gain'' is actually a collapse once human text is included. Across two corpora and nine detectors, we find that reported gains under attack vanish under two-sided evaluation, and the cost of ignoring the human side can be devastating.

Our findings also show that some damaging attacks are surprisingly easy to undo. Simple preprocessing steps, such as stripping zero-width characters, collapsing repeated spaces, or folding homoglyphs restore detection performance to near-clean levels. By contrast, word-level synonym substitution remains effective, meaning-preserving, and irreversible.

Our main contributions are as follows:

\begin{itemize}[leftmargin=1.2em,itemsep=2pt,topsep=3pt]
    \item We introduce a \textbf{two-sided protocol} that applies each attack to human as well as machine text and scores the two against each other.

    \item We show that \textbf{reported gains under attack are not real}. Across two corpora and nine detectors, 39 of 162 cells show an AUROC gain, 31 significantly, and none survives two-sided as a significant gain.
    
    \item We measure \textbf{the cost of the missing half}. At a threshold set to 1\% FPR of clean human documents, attacked human text is flagged at up to 100\%.
    
    \item We show that \textbf{DIPPER paraphrase no longer evades} most detectors. It raises AUROC for five of nine while lowering detection for all nine, and a defense built on it fails for the same reason.
    
    \item We test \textbf{sanitization as a defense}. Zero-width and whitespace attacks are fully recoverable, and homoglyphs only with a matching table. Word substitutions are not recoverable.
    
    \item We show that \textbf{generator choice and text length} move detectors further than any attack, and that short text is where false accusations concentrate.

\end{itemize}

The source code and data will be made publicly available upon acceptance.

\section{Related Work}
\paragraph{Zero-shot and supervised detection.}
Likelihood-based statistics date to GLTR \citep{gehrmann2019gltr}. DetectGPT
scores how far a document's likelihood drops when a few words are changed
\citep{mitchell2023detectgpt}, and Fast-DetectGPT computes the same thing in
one forward pass \citep{bao2024fastdetectgpt}. Binoculars compares a pair of
models \citep{hans2024binoculars}, and Lastde reads token likelihoods as a
time series \citep{xu2025lastde}. Supervised detectors instead fine-tune
encoders on labeled data \citep{guo2023hc3}, and RADAR trains one against a
paraphraser \citep{hu2023radar}.

\paragraph{Attacks and robustness benchmarks.}
\citet{krishna2023paraphrasing} established paraphrase evasion and proposed
retrieval as a defense. RAID \citep{dugan2024raid} provides the largest
attack-annotated corpus, including the character-level attacks we study.
DetectRL \citep{wu2024detectrl} is the closest prior work. It does attack
human text. But it scores that text against \emph{clean} machine text, which
is the usual one-sided design in mirror image. It also reports only AUROC and F1, with the threshold chosen by Youden's $J$ on the evaluation set itself. Because that threshold is refitted for every condition, it cannot show a change in the false-positive rate. \citet{creo2025silverspeak} show that homoglyph substitution drives seven detectors to near-random behavior and name unicode normalization as a possible defense without evaluating it. RAFT \citep{wang2024raft} and GREATER \citep{greater2025} establish word-level substitution as an attack family where \citet{saiteja2025modeling} organise attacks into a taxonomy, report TPR at strict FPR, and build a detector that normalizes its input and feeds the discrepancy between the raw and canonical text to a classifier. That treats normalization as a feature source inside a trained model. We instead measure it as a preprocessing defense in isolation, asking how much of a detector's
clean operating point plain sanitization restores (Section~\ref{sec:defense-works}), which their design does not report.

Paraphrase \emph{does} still defeat detectors when the paraphraser is
purpose-built for evasion: \citet{shportko2025resilience} drive Binoculars
from $0.750$ to $0.553$ F1 with a commercial humanizer,
\citet{ranganath2026stealthrl} reach mean TPR@1\%FPR $0.001$ with a
reinforcement-trained policy, and \citet{zha2025padben} report failure under
iterated paraphrase. Our result concerns DIPPER,  the paraphraser of \citet{krishna2023paraphrasing}, applied once. What has expired is not paraphrasing but \emph{incidental} paraphrasing.

\paragraph{Attacks that help the detector.}
\citet{riverasoto2026fingerprints} show that style transfer can \emph{raise}
detectability. The rewriting leaves its own signature, and Binoculars goes
from $79$ to $92$ AUROC. We see the same effect for DIPPER paraphrase and for unicode attacks on likelihood detectors, and we show the gain comes from how it is measured. \citet{lacava2026luminol} independently report that
Binoculars and Fast-DetectGPT break under homoglyph and zero-width attacks.
\citet{mcgovern2025fingerprints} find that generator fingerprints survive
paraphrase, which our results support.

\paragraph{False positives on human text.} \citet{saha2025polished} are the
closest prior work. They run human text through an LLM and show detectors
flag it, with GLTR rising from $6.8\%$ to $40.9\%$ on lightly edited text.
Three things separate our result. Their edits are rewrites a reader can see,
where ours look identical. Their threshold is tuned for accuracy, where ours
is fixed to flag $1\%$ of held-out clean human text. And they never attack
the machine text, so the two sides are not treated alike.
\citet{liang2023bias} show detectors misclassify non-native English writing,
with false-positive rates above $60\%$ on TOEFL essays.

\paragraph{Paraphrase defenses.} \citet{tufts2025practical} also use strict
FPR thresholds, but only on unattacked human text. \citet{riverasoto2025inversion} train a model to undo paraphrasing and gain $+22\%$ AUROC. But our experiment shows an opposite outcome (Section~\ref{sec:defense-paraphrase}).

\section{Experimental Setup}
\label{sec:setup}

\paragraph{Data.} We use two datasets: RAID\footnote{\url{https://github.com/liamdugan/raid}}\citep{dugan2024raid} and DetectRL\footnote{\url{https://github.com/NLP2CT/DetectRL}} \citep{wu2024detectrl}, and Figure~\ref{fig:pipeline} shows how both are sampled and scored. From RAID, we sample evenly across eight domains, selecting 90 machine and 180 human documents per domain of at least 30 words, each with eleven attack variants. This yields 27{,}029 texts from 2{,}160 sources, spanning 11 generators with greedy and sampling decoding. For length analysis, 20\% of sources are truncated to 50, 100, and 200 words. Human texts receive the same attacks as machine texts, enabling false-positive analysis.

\noindent DetectRL is sampled similarly: 1{,}440 human and 720 machine sources across four domains and four generators, each with six attack variants. Since DetectRL lacks zero-width attacks, we add RAID’s implementation (Section~\ref{sec:replication}), producing 17{,}280 texts. We use its raw corpus rather than prepared splits, as two splits mislabel 33\% of human text as machine (Appendix~\ref{sec:detectrl-detail}).

\paragraph{Detectors.} We evaluate nine detectors, all normalized so that higher scores indicate more machine-like text. They fall into three families:  
\textbf{Modern zero-shot}: Binoculars (Falcon-7B pair), Fast-DetectGPT (analytic, on gpt-neo-2.7B, gpt2-xl, gpt2), and Lastde (gpt-neo-2.7B).  
\textbf{Legacy likelihood}: mean log-likelihood and log-rank (gpt-neo-2.7B).  
\textbf{Trained classifiers}: RADAR \citep{hu2023radar} and HC3-trained RoBERTa \citep{guo2023hc3}. The detector list is available at Table~\ref{tab:detectors}.

\paragraph{Metrics and Protocol.} For each detector and attack we report two key metrics:  
\textbf{AUROC} (Area Under the Receiver Operating Characteristic), which measures the probability that a detector ranks a machine text higher than a human text, and  
\textbf{TPR} (True Positive Rate) at $1\%$ FPR (False Positive Rate), which measures the fraction of machine texts correctly flagged when only $1\%$ of clean human texts are falsely flagged, following \citet{krishna2023paraphrasing}.

Each threshold is the 99th percentile of the scores on half the clean human documents, and everything is measured on the other half (Appendix~\ref{sec:calibration}). We also report two numbers the one-sided protocol never produces: the share of attacked human documents that the same thresholds flag as AI and the AUROC between the two attacked classes. Let $S$ for a detector's score, $H$ and $M$ for clean human and machine text, and $H_a$, $M_a$ for the same documents under attack $a$, the two AUROCs differ in one argument:
\begin{align}
\mathrm{AUROC}_{1}(a) &= \mathrm{AUROC}\!\left(S(H),\; S(M_a)\right)\\
\mathrm{AUROC}_{2}(a) &= \mathrm{AUROC}\!\left(S(H_a),\, S(M_a)\right)
\end{align}
Every comparison carries a $95\%$ bootstrap confidence interval (CI) from $1{,}000$ seeded replicates, each refitting its own threshold. On the
held-out half, the thresholds flag $0.011$ of clean human documents on RAID and $0.009$ on DetectRL against a target of $0.01$, pooled over the
nine detectors (Table~\ref{tab:replication}), so the false accusations of Section~\ref{sec:twosided} come from the attacks and not from a badly
fitted threshold.

\paragraph{Length control and excluded data.} For the length analysis, we cut human text to the same lengths as the truncated AI text so that AUROC compares machine against human rather than short against long. We dropped one document, a paraphrased text containing nothing but an
ellipsis that Lastde could not score, leaving $243{,}254$ scores.

\section{Results}
\label{sec:results}
\begin{figure}[t]
  \centering
  \includegraphics[width=\columnwidth]{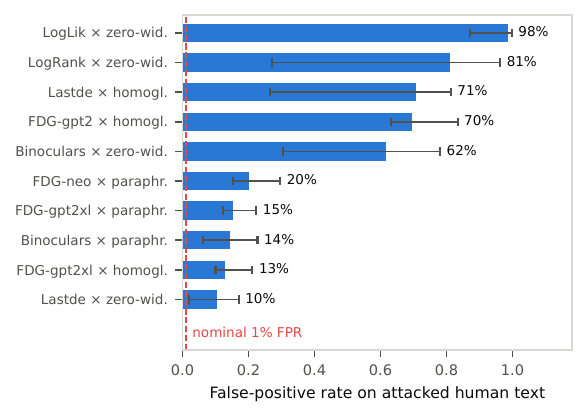}
  \caption{False-positive rate on \emph{attacked human} text at thresholds
  calibrated for 1\% FPR on clean human text (red line).}
  \label{fig:fpr}
\end{figure}

\begin{figure*}[!t]
  \centering
  \begin{minipage}[b]{0.48\textwidth}
    \centering
    \includegraphics[width=\linewidth]{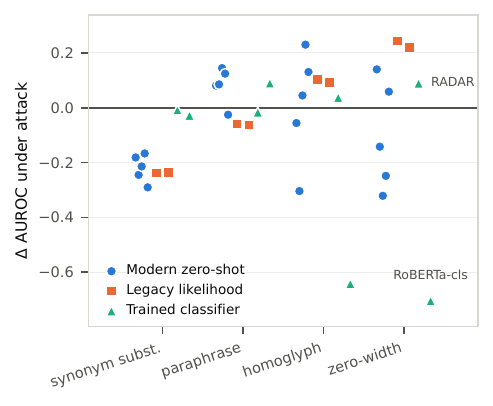}
    \caption{Change in AUROC under attack, one mark per detector by family.
    Synonym substitution breaks probability-based detectors, Unicode
    attacks break the classifiers.}
    \label{fig:inversion}
  \end{minipage}\hfill
  \begin{minipage}[b]{0.48\textwidth}
    \centering
    \includegraphics[width=\linewidth]{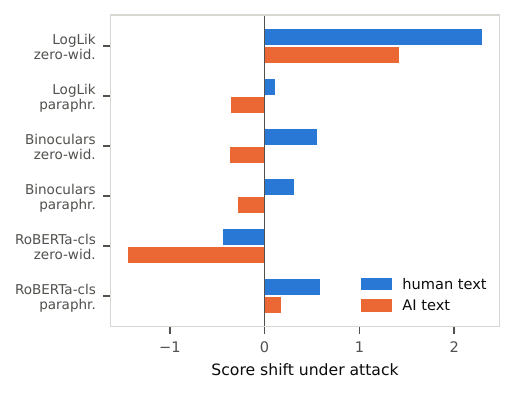}
    \caption{Paired score shift under attack for human (blue) and AI
    (orange) text, in units of each detector's clean human to AI gap.
    Positive means toward ``AI'', negative toward ``human''.}
    \label{fig:twosided}
  \end{minipage}
\end{figure*}

We evaluate nine detectors across eleven RAID attacks and repeat the study on DetectRL  (Table~\ref{tab:replication}, Figure~\ref{fig:pipeline}). Importantly, both human and machine texts are attacked. When scored against clean human text, 22 detector–attack pairs appear to improve. However, when scored against attacked human text, 39 pairs become significantly worse (Table~\ref{tab:replication}, Table~\ref{tab:twosided}). These apparent improvements occur because attacks also push human text toward ``AI-like'' (Figure~\ref{fig:twosided}), causing detectors to misclassify up to $98\%$ of innocent writers at thresholds calibrated to flag only $1\%$ (Figure~\ref{fig:fpr}, Table~\ref{tab:fprgrid}).

\begin{figure*}[t]
  \centering
  \includegraphics[width=\textwidth]{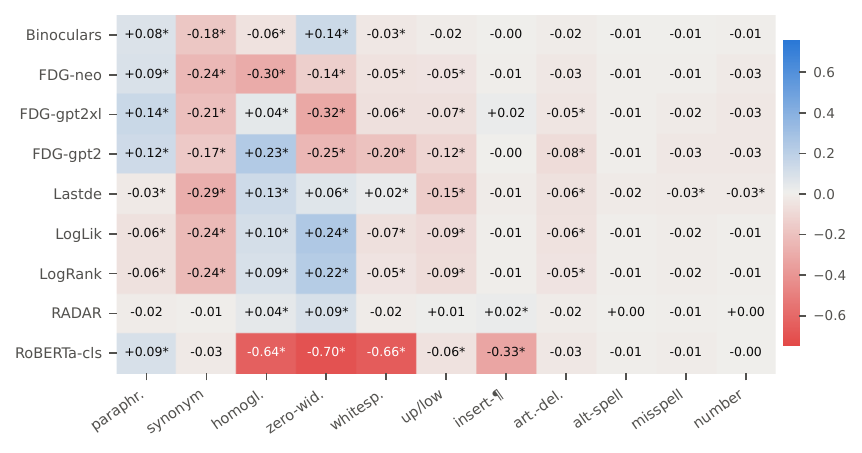}
  \caption{Effect of each attack on each detector ($\Delta$AUROC vs.\ no
  attack). Blue helps the detector; red breaks it; $*$ marks deltas whose
  95\% bootstrap CI excludes zero. Effects are family-antagonistic: the
  zero-width column simultaneously contains the largest positive and largest
  negative entries.}
  \label{fig:heatmap}
\end{figure*}

We further find that \textit{paraphrase} no longer reliably hides AI text. AUROC even rises for five of nine detectors, yet all nine detect fewer AI documents at a fixed cutoff (Table~\ref{tab:paraphrase}, Table~\ref{tab:tprgrid}). By contrast, simple character edits cause far greater damage (Figure~\ref{fig:heatmap}). Preprocessing can undo two of the three worst attacks, but word substitution remains resistant (Table~\ref{tab:variants}, Table~\ref{tab:synonym}). Finally, for eight of the nine detectors, the choice of generator influences performance more strongly than any attack.

\subsection{Paraphrasing no longer evades most detectors}
\label{sec:paraphrase}

DIPPER paraphrase \emph{raises} AUROC for five of the nine detectors, yet all of them catch fewer AI texts at a fixed cutoff (Table~\ref{tab:paraphrase}). The attack shifts human text toward ``AI-like'' while making AI scores more uniform (Figure~\ref{fig:twosided}). In effect, the paraphraser leaves its own signature on the text rather than removing the generator’s. Binoculars illustrates this most clearly: AUROC increases by $0.081$, but detection falls from $0.74$ to $0.68$ (CI $[-0.129,-0.008]$), with both changes statistically significant (Figure~\ref{fig:divergence}).

\begin{table}[t]
\centering\small
\begin{tabular}{@{}lcccc@{}}
\toprule
& \multicolumn{2}{c}{AUROC} & \multicolumn{2}{c}{TPR@1\%FPR} \\
\cmidrule(lr){2-3}\cmidrule(l){4-5}
Detector & clean & para. & clean & para. \\
\midrule
Binoculars   & .854 & \textbf{.935} & .74 & .68 \\
FDG (neo)    & .828 & \textbf{.913} & .69 & .59 \\
FDG (gpt2-xl)& .763 & \textbf{.907} & .58 & .53 \\
FDG (gpt2)   & .725 & \textbf{.849} & .38 & .26 \\
Lastde       & .840 & .814 & .41 & \textbf{.18} \\
LogLik       & .752 & .693 & .21 & \textbf{.06} \\
LogRank      & .771 & .708 & .23 & \textbf{.07} \\
RADAR        & .874 & .857 & .24 & .21 \\
RoBERTa-cls  & .724 & \textbf{.814} & .23 & .19 \\
\bottomrule
\end{tabular}
\caption{Clean vs.\ paraphrase-attacked performance.  Bold marks where AUROC rises under attack while detection at $1\%$ FPR falls.}
\label{tab:paraphrase}
\end{table}

\subsection{The simplest attacks do the most damage}
Figure~\ref{fig:heatmap} maps all 99 detector–attack combinations. The most damaging attacks are simple character edits. For example, zero-width insertion reduces RoBERTa by $0.705$ AUROC and drives its detection rate to zero. Homoglyph substitutions show mixed effects: they \emph{raise} Fast-DetectGPT at GPT-2 by $+0.230$, but lower it by $0.304$ at GPT-Neo. No single attack harms all nine detectors, and the same attack can benefit one detector while severely weakening another, depending on the backbone used  (Figure~\ref{fig:inversion}).

\subsection{Attacks move both classes}
\label{sec:twosided}
We score attacked human text against attacked AI text and measure the score shift. 

For a class $C$ under attack $a$, the shift is defined as the paired score change divided by the clean gap between the classes:
\begin{equation}
\Delta_a(C) = \frac{1}{|C|}\sum_{x \in C}
  \frac{S(x_a) - S(x)}{\bar{S}(M) - \bar{S}(H)}
\end{equation}

For example, zero-width insertion moves human text $+2.3$ separation units toward ``AI,'' while machine text shifts only $+1.4$ (Figure~\ref{fig:twosided}). This causes innocent human text to be flagged. Under log-likelihood, zero-width insertion raises AUROC from $0.752$ (clean) to $0.996$ (one-sided), but only $0.576$ (two-sided). At a threshold set to flag $1\%$ of clean human text, attacked human text is flagged $98\%$ of the time (95\% CI $[0.87,1.00]$). The same attack produces $81\%$ false positives for log-rank and $62\%$ for Binoculars (Figure~\ref{fig:fpr}).

\paragraph{A stronger detector is a worse accuser.} Fast-DetectGPT's reference implementation defaults to a falcon-7b pair, so we re-ran it there too (Table~\ref{tab:backbone}). It performed better on clean text by increasing AUROC for both RAID and DetectRL. But it flags $99.9\%$ of attacked human documents, whereas the weaker backbone flagged only $2\%$. So, improving the detector made it confidently wrong about more innocent writers.


\begin{table}[t]
\centering\small
\setlength{\tabcolsep}{3pt}
\begin{tabular}{@{}lccccccc@{}}
\toprule
& clean & \multicolumn{3}{c}{zero-width} & \multicolumn{3}{c}{homoglyph} \\
\cmidrule(lr){3-5}\cmidrule(l){6-8}
backbone & AUROC & 1-s & 2-s & FPR & 1-s & 2-s & FPR \\
\midrule
\multicolumn{8}{@{}l}{\emph{RAID}} \\
\hspace{4pt}gpt-neo & 0.828 & 0.69 & 0.64$^{\dagger}$ & 0.02 & 0.52 & 0.60$^{\dagger}$ & 0.01 \\
\hspace{4pt}falcon-7b & 0.854 & 1.00$^{*}$ & 0.58$^{\dagger}$ & 1.00 & 0.84 & 0.64$^{\dagger}$ & 0.23 \\
\addlinespace[2pt]
\multicolumn{8}{@{}l}{\emph{DetectRL}} \\
\hspace{4pt}gpt-neo & 0.902 & 0.67 & 0.66$^{\dagger}$ & 0.01 & -- & -- & -- \\
\hspace{4pt}falcon-7b & 0.942 & 1.00$^{*}$ & 0.47$^{\dagger}$ & 0.99 & -- & -- & -- \\
\bottomrule
\end{tabular}
\caption{Fast-DetectGPT at gpt-neo-2.7B and at the falcon-7b pair authors recommend. \emph{1-s} and \emph{2-s} are one- and two-sided AUROC, and \emph{FPR} is the fraction of attacked human documents flagged. $^{*}$ marks a significant gain over clean, $^{\dagger}$ a significant
loss. DetectRL's homoglyph attack also substitutes words, so they are left blank.}
\label{tab:backbone}
\end{table}

\subsection{Replication of results on DetectRL corpus}
\label{sec:replication}

We repeat the study on DetectRL \citep{wu2024detectrl}, which differs from RAID in domain (arXiv, news, stories, reviews), generator (GPT-3.5-turbo, PaLM-2, Claude-instant, Llama-2-70b), and attack implementation. Six of its seven attacks are unique to DetectRL. Since DetectRL does not include a zero-width attack, we generate one from its documents using RAID's insertion rule.

The false accusations are even worse here than on RAID. Under zero-width insertion, log-likelihood and log-rank flag \emph{all} $1{,}440$ attacked human documents, whereas Lastde flags $68\%$ and Binoculars $31\%$. The overall pattern mirrors RAID: out of 63 cells, none shows a significant two-sided gain (Table~\ref{tab:replication}). Under back-translation, RADAR’s AUROC rises from $0.921$ to $0.956$, but when measured two-sided it falls to $0.554$, and the detector wrongly flags $37\%$ of human documents (Table~\ref{tab:fprgrid-dr}).

\begin{table}[t]
\centering\small
\begin{tabular}{@{}lcc@{}}
\toprule
& RAID & DetectRL \\
\midrule
domains & 8 & 4 \\
generator configurations & 22 & 4 \\
(detector, attack) cells & 99 & 63 \\
\midrule
realized FPR, clean human & 0.011 & 0.009 \\
worst FPR, attacked human & 0.98 & 1.00 \\
\midrule
one-sided AUROC gains & 22 & 17 \\
\quad significant & 18 & 13 \\
\textbf{two-sided gains} & \textbf{0} & \textbf{0} \\
significant two-sided losses & 39 & 54 \\
\bottomrule
\end{tabular}
\caption{The two-sided result on both corpora}
\label{tab:replication}
\end{table}

\subsection{Short text draws false accusations}
\label{sec:length}

Fast-DetectGPT struggles more on short documents. Its true positive rate at a $1\%$ threshold drops from $0.72$ to $0.32$ when the text length is reduced to 50 words, while its false-accusation rate remains at $0.01$. Binoculars remains stable across all lengths. RADAR appears to improve, rising from $0.24$ to $0.74$, but in reality it flags $61\%$ of 50-word \emph{human} texts at the same threshold (Figure~\ref{fig:length}).

Short texts such as chat messages, exam answers, and social posts are where false accusations concentrate. In these cases, the choice of detector largely determines the outcome. Thus, short texts amplify detector bias and increase the risk of misclassification.

\subsection{Only synonym substitution survives sanitization}
\label{sec:synonym}

\begin{table*}[t]
\centering\small
\setlength{\tabcolsep}{6pt}
\begin{tabular}{@{}llccccc@{}}
\toprule
& & & \multicolumn{2}{c}{TPR@1\%} & \multicolumn{2}{c}{score shift} \\
\cmidrule(lr){4-5}\cmidrule(l){6-7}
detector & family & $\Delta$AUROC & clean & syn. & AI & human \\
\midrule
Binoculars & modern zero-shot & -0.181 & 0.74 & 0.30 & -0.74 & -0.30 \\
FDG-neo & modern zero-shot & -0.245 & 0.69 & 0.22 & -1.01 & -0.68 \\
FDG-gpt2xl & modern zero-shot & -0.214 & 0.58 & 0.23 & -1.14 & -0.74 \\
FDG-gpt2 & modern zero-shot & -0.167 & 0.38 & 0.11 & -1.10 & -0.74 \\
Lastde & modern zero-shot & -0.290 & 0.41 & 0.09 & -0.78 & -0.38 \\
LogLik & legacy stat. & -0.238 & 0.21 & 0.09 & -0.93 & -0.77 \\
LogRank & legacy stat. & -0.236 & 0.23 & 0.09 & -0.90 & -0.73 \\
RADAR & trained clf. & -0.008 & 0.24 & 0.16 & -0.09 & +0.01 \\
RoBERTa & trained clf. & -0.028 & 0.23 & 0.17 & -0.14 & -0.05 \\
\bottomrule
\end{tabular}
\caption{Synonym substitution by detector family. Probability-based
detectors collapse; trained classifiers are nearly unaffected. Score shifts
are paired (attacked minus source) in units of each detector's clean
human--AI separation: the attack pushes both classes toward ``human'', but
AI text roughly twice as far.}
\label{tab:synonym}
\end{table*}

Synonym substitution replaces about a fifth of the words in a document (median 21\%). It keeps the meaning better than paraphrase, at $0.92$ sentence similarity against $0.87$ (Table~\ref{tab:semantic}). 

It is most effective against probability-based detectors. It drops Binoculars by $0.181$ AUROC and from $0.74$ to $0.30$ TPR@1\%FPR, whereas Lastde and legacy likelihood pairs fall by $0.290$ and $0.24$, respectively. The attack pushes both classes toward ``human'', but machine text more than twice as far (Table~\ref{tab:synonym}). Machine text is built from high-probability words, and replacing those with less likely synonyms costs it the most.

It barely has any effect on the trained classifiers. Because they read word choice and style rather than token probabilities, and a synonym word leaves those intact. For zero-shot detectors, its False-positive rate on attacked human text is ${\leq}0.003$ and acts like pure evasion.

\subsection{Sanitization removes the Unicode attacks entirely}
\label{sec:defense-works}
Homoglyphs and zero-width characters change the tokens a detector sees while leaving the page identical to a reader, so they can be reversed. We remove the zero-width characters and use a curated table of 47 pairs (Table~\ref{tab:curated}) to replace each look-alike character with the letter it imitates, so Cyrillic \textsc{u+0430} becomes Latin \texttt{a}. We then apply NFKC normalization to turn text back into its plain form. Each step repairs only its own kind of change, which is why all three steps are needed.

\begin{figure}[t]
  \centering
  \includegraphics[width=\columnwidth]{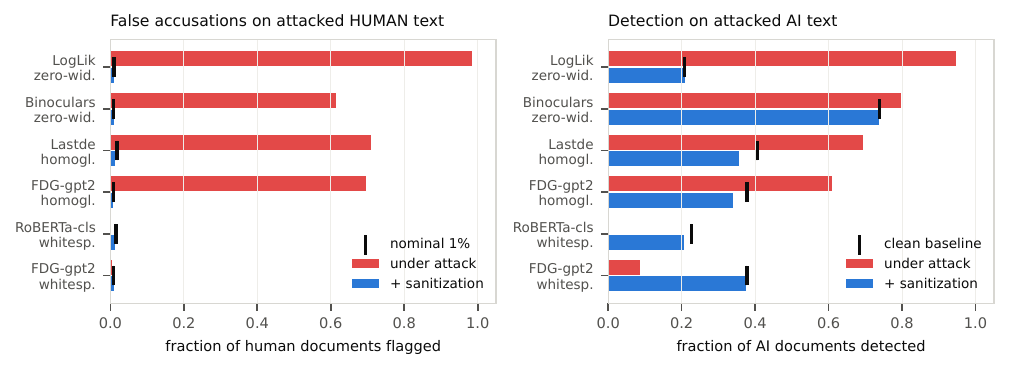}
  \caption{Attacked documents re-scored after sanitization, which strips zero-width characters, folds confusables, collapses whitespace, and applies NFKC. Thresholds are unchanged from the clean calibration.}
  \label{fig:defense}
\end{figure}

\begin{table}[t]
\centering\small
\begin{tabular}{@{}lccc@{}}
\toprule
& \multicolumn{2}{c}{trigram sim.\ to source} & exact \\
\cmidrule(lr){2-3}
attack & attacked & sanitized & restore \\
\midrule
zero-w. & 0.000 & 0.999 & 97\% \\
homogl. & 0.118 & 0.988 & 31\% \\
synonym & 0.689 & 0.688 & 0\% \\
paraphr. & 0.504 & 0.504 & 0\% \\
whitesp. & 0.930 & 0.929 & 0\% \\
up/low & 0.948 & 0.947 & 0\% \\
art-del & 0.987 & 0.987 & 3\% \\
ins-par & 0.974 & 0.974 & 4\% \\
number & 0.979 & 0.979 & 47\% \\
alt-sp. & 0.993 & 0.993 & 34\% \\
misspell & 0.991 & 0.991 & 22\% \\
\bottomrule
\end{tabular}
\caption{Text similarity to the source before and after sanitization, and the share of documents restored exactly. For word-level attacks, a non-zero exact restore means the attack found nothing to change, not that sanitization repaired it.}
\label{tab:fidelity}
\end{table}

\paragraph{The text and the detection both come back.} Zero-width insertion leaves $0.000$ character trigram similarity to the source, and sanitization brings it to $0.999$, with $97\%$ of documents restored byte for byte (Table~\ref{tab:fidelity}). The remaining $3\%$ differ only because NFKC also normalizes characters the source already contained. Re-scored at the unchanged thresholds, every detector returns to its clean operating point (Figure~\ref{fig:defense}). Log-likelihood's false-accusation rate under zero-width falls from $0.985$ back to $0.010$, and homoglyph detection recovers to $86$ to $104\%$ of clean. DetectRL's homoglyph attack also swaps words, so recovery there stops at $16$ to $91\%$.

\paragraph{Extra spaces are easy to undo.} The whitespace attack adds extra spaces between words. So replacing every group of spaces with a single space brings detection back to $90$ to $100\%$ of its clean value for all nine detectors (Table~\ref{tab:variants}).

\begin{table}[t]
\centering\small
\begin{tabular}{@{}lccc@{}}
\toprule
& \multicolumn{3}{c}{detection after cleaning, \% of clean} \\
\cmidrule(l){2-4}
attack & curated & full UTS \#39 & curated $+$ ws \\
\midrule
zero-w. & 100--100 & 100--102 & 99--100 \\
homogl. & 86--104 & 80--105 & 86--103 \\
whitesp. & 0--92 & 92--101 & 90--100 \\
\midrule
\emph{cost} & 0/9 & 4/9 & 1/9 \\
\bottomrule
\end{tabular}
\caption{Detection after each sanitizer configuration, as a percentage of
clean detection. Cells show the range over the nine detectors. \emph{Cost}
counts detectors that wrongly flag more clean human text.}
\label{tab:variants}
\end{table}

\paragraph{A larger confusable table is worse.} Unicode's full UTS~\#39 list removes more of the substituted characters than our curated table ($98.0\%$ against $61.8\%$, Table~\ref{tab:confusables-tradeoff}), yet it does not recover detection and raises false accusations on clean human text for four of nine detectors, where the curated table raises none (Table~\ref{tab:variants}). The large table maps many different characters to the same letter. For example, a look-alike capital \texttt{I} comes back as a lowercase \texttt{l}. On RAID, it rewrites $15\%$ of clean documents against the curated table's $3\%$ (Table~\ref{tab:confusables-tradeoff}). Unicode attacks look strong in the literature when the text is not cleaned, so the reported number measures whether the text was processed, not how good the detector is. One line of preprocessing decides whether these attacks look devastating or harmless.

\subsection{Generator choice matters more than any attack}
\label{sec:generators}
We score each detector on clean text from eleven generators, each under greedy decoding and under sampling, 22 setups in all (Table~\ref{tab:generators}). Binoculars separates ChatGPT text from human text perfectly and scores $1.000$ AUROC, but on MPT with sampling, it scores only $0.430$.  That gap of $0.570$ is three times the damage its worst attack does ($0.181$). The gap is $0.374$ for RADAR and $0.644$ for log-likelihood, and only RoBERTa's zero-width loss $0.705$
exceeds its generator gap. The weak setups are older base models with sampling, and every perfect score uses greedy decoding. A study that evaluates only chat models is not likely to have encountered the ones that defeat the detector.

\subsection{Paraphrasing as a defense}
\label{sec:defense-paraphrase}

Section~\ref{sec:paraphrase} showed that paraphrasing makes text appear more machine-like. This suggests a possible defense: rather than scoring the document directly, paraphrase it first and then score the output. An attacker can craft a document to slip past a detector, but cannot control what a paraphraser will produce. To test this, we sampled 100 human and 100 machine documents across RAID’s eight domains, along with those same machine documents under paraphrase attack, and generated $K{=}5$ paraphrases of each.


Under Fast-DetectGPT, machine text fell from a mean score of $4.56$ to $-0.50$, while human text fell only from $-0.24$ to $-1.68$. The gap between the classes shrank from $4.80$ to $1.18$. As a result, the five paraphrases separate the classes \emph{worse} than the single document they replaced, with AUROC dropping from $0.95$ to $0.81$ on clean text and from $0.96$ to $0.77$ on attacked text. Log-likelihood shows the same trend, falling from $0.86$ to $0.63$.

A paraphraser’s output reflects the paraphraser itself, not the input. Paraphrases overwrite the authorship signal of the text, which determines what the detector reads. The paraphrases still separate the classes at $0.81$, consistent with \citet{mcgovern2025fingerprints}, but this is worse than the $0.95$ separation the defense was meant to improve.

\section{Recommendations}
\textbf{(1) Attack the human text too.} Report attacked-human FPR alongside attacked-AI TPR. Attacking only AI text overestimates robustness.
\textbf{(2) Report TPR at a fixed FPR, with CIs.} AUROC can rise while detection at a $1\%$ threshold falls (Figure~\ref{fig:divergence}).
\textbf{(3) Report the generator mix.} It moves performance further than any attack does (Section~\ref{sec:generators}), so a number quoted without it is not comparable across papers.
\textbf{(4) Report results by text length.} A detector can look strongest on short text while accusing most of its short human inputs (Section~\ref{sec:length}).
\textbf{(5) Report per-family results.} Robustness does not transfer between families, so a claim about one detector says nothing about the other. 
\textbf{(6) Strip zero-width characters and collapse whitespace.} Both are free and both fully reverse their attacks (Section~\ref{sec:defense-works}). Fold confusables with a table matched to the attacks you expect. And say whether you sanitized, because otherwise results for these attacks measure preprocessing rather than detectors.
\textbf{(7) Study the attacks that survive sanitization.} Word-level substitution keeps the meaning, cannot be undone, and costs Binoculars more
than any other attack, yet the main benchmarks omit it.

\section{Conclusion}
\label{sec:conclusion}

Our results show that current AI text detectors are systematically fragile. Simple character-level edits, paraphrasing, and short text lengths distort both human and machine classes, producing misleading AUROC gains while inflating false-positive rates. Replication across RAID and DetectRL confirms that these weaknesses are structural rather than dataset-specific. Stronger backbones can make a detector more confident but more wrong, and defenses such as paraphrasing overwrite authorship signals rather than preserving them.

These findings highlight the need for transparent evaluation protocols. Robustness must be measured two-sided, with attacked-human false positives reported alongside attacked-AI true positives, and results disaggregated by generator, text length, and detector family. While sanitization can reverse the Unicode attacks, meaning-preserving substitutions remain a critical open challenge. In sum, AI text detection is not yet reliable for high-stakes use, and progress will depend on careful reporting and a focus on attacks that survive preprocessing.

\section{Limitations}
Our corpora are English-only and were collected in 2024, so newer generators
may shift the absolute numbers, though the two-sided result reproduces across both. We test post-hoc text attacks, not adversarial prompting
\citep{tufts2025practical} or fine-tuned evasion, and rewriting defenses, not retrieval \citep{krishna2023paraphrasing} or watermarking. Some DetectRL machine documents open with a chat-style introduction, mostly
from one generator. They are easy to detect. But human text never opens that way, so the false-accusation results hold.

\section{Ethical Considerations}
\label{sec:ethics}

This study relies exclusively on datasets that are publicly available, ensuring compliance with accepted research standards. No personally identifiable information (PII) was collected, stored, or analyzed at any stage of the work. All experiments were conducted on text corpora drawn from open sources, and no private or sensitive data were involved.

The findings highlight limitations in current AI text detectors, but do not introduce new risks. We emphasize that any models, code, or evaluation protocols released should be applied responsibly. In particular, detectors should not be used for punitive or surveillance purposes against individuals, and their deployment must align with responsible AI guidelines. Researchers and practitioners are encouraged to avoid harmful or misleading applications and to ensure that these tools are used only in socially beneficial contexts.

\bibliography{custom}
\appendix

\setcounter{dbltopnumber}{3}
\setcounter{topnumber}{4}
\setcounter{totalnumber}{6}
\renewcommand{\dbltopfraction}{0.95}
\renewcommand{\topfraction}{0.95}
\renewcommand{\textfraction}{0.02}
\renewcommand{\dblfloatpagefraction}{0.85}
\renewcommand{\floatpagefraction}{0.85}

\section{RAID full result grid}
\label{sec:grids}

Table~\ref{tab:twosided} reports the two-sided AUROC from Section~\ref{sec:twosided} where attacked human text scored against attacked AI text. Figure~\ref{fig:heatmap} plots the one-sided values for the same grid, with significance markers. Reading them together shows what the one-sided protocol hides. The pairs that look like the largest gains one-sided are among the largest losses two-sided.

Table~\ref{tab:tprgrid} reports detection at the $1\%$ threshold, and Table~\ref{tab:fprgrid} the false-accusation rate on attacked human text for every detector and attack. Both numbers are needed. Because one of them can look fine while the other is catastrophic. Table~\ref{tab:generators} breaks the clean AUROC out by generator, which Section~\ref{sec:generators} summarises. Every detector's best setup is a greedy decode and every worst is a sampled one.

\begin{table*}[t]
\centering\footnotesize
\setlength{\tabcolsep}{2.0pt}
\begin{tabular}{@{}llcccccccccccc@{}}
\toprule
& \multicolumn{13}{c}{AUROC} \\
\cmidrule(l){2-14}
detector & clean & protocol & paraphr. & synonym & homogl. & zero-w. & whitesp. & up/low & ins-par & art-del & alt-sp. & misspell & number \\
\midrule
Binoculars & 0.85 & 1-sided & 0.93$^{*}$ & 0.67 & 0.80 & 0.99$^{*}$ & 0.82 & 0.84 & 0.85 & 0.84 & 0.85 & 0.85 & 0.85 \\
& & 2-sided & 0.81$^{\dagger}$ & 0.80$^{\dagger}$ & 0.68$^{\dagger}$ & 0.65$^{\dagger}$ & 0.84 & 0.86 & 0.85 & 0.86 & 0.85 & 0.85 & 0.86 \\
\addlinespace[2pt]
FDG-neo & 0.83 & 1-sided & 0.91$^{*}$ & 0.58 & 0.52 & 0.69 & 0.77 & 0.78 & 0.82 & 0.80 & 0.82 & 0.82 & 0.80 \\
& & 2-sided & 0.73$^{\dagger}$ & 0.77$^{\dagger}$ & 0.60$^{\dagger}$ & 0.64$^{\dagger}$ & 0.81 & 0.82 & 0.81 & 0.83 & 0.83 & 0.82 & 0.82 \\
\addlinespace[2pt]
FDG-gpt2xl & 0.76 & 1-sided & 0.91$^{*}$ & 0.55 & 0.81$^{*}$ & 0.44 & 0.70 & 0.70 & 0.78 & 0.72 & 0.75 & 0.74 & 0.73 \\
& & 2-sided & 0.74 & 0.69$^{\dagger}$ & 0.52$^{\dagger}$ & 0.58$^{\dagger}$ & 0.71$^{\dagger}$ & 0.74 & 0.80 & 0.76 & 0.76 & 0.76 & 0.75 \\
\addlinespace[2pt]
FDG-gpt2 & 0.72 & 1-sided & 0.85$^{*}$ & 0.56 & 0.96$^{*}$ & 0.48 & 0.52 & 0.61 & 0.72 & 0.64 & 0.71 & 0.69 & 0.69 \\
& & 2-sided & 0.66$^{\dagger}$ & 0.68$^{\dagger}$ & 0.45$^{\dagger}$ & 0.58$^{\dagger}$ & 0.57$^{\dagger}$ & 0.68$^{\dagger}$ & 0.75 & 0.72 & 0.72 & 0.71 & 0.71 \\
\addlinespace[2pt]
Lastde & 0.84 & 1-sided & 0.81 & 0.55 & 0.97$^{*}$ & 0.90$^{*}$ & 0.86$^{*}$ & 0.69 & 0.83 & 0.78 & 0.82 & 0.81 & 0.81 \\
& & 2-sided & 0.73$^{\dagger}$ & 0.75$^{\dagger}$ & 0.55$^{\dagger}$ & 0.55$^{\dagger}$ & 0.82 & 0.81 & 0.82 & 0.82 & 0.83 & 0.83 & 0.84 \\
\addlinespace[2pt]
LogLik & 0.75 & 1-sided & 0.69 & 0.51 & 0.86$^{*}$ & 1.00$^{*}$ & 0.68 & 0.66 & 0.74 & 0.70 & 0.74 & 0.74 & 0.74 \\
& & 2-sided & 0.67$^{\dagger}$ & 0.73 & 0.62$^{\dagger}$ & 0.58$^{\dagger}$ & 0.74 & 0.75 & 0.74 & 0.75 & 0.75 & 0.75 & 0.77 \\
\addlinespace[2pt]
LogRank & 0.77 & 1-sided & 0.71 & 0.54 & 0.86$^{*}$ & 0.99$^{*}$ & 0.72 & 0.68 & 0.76 & 0.72 & 0.76 & 0.76 & 0.76 \\
& & 2-sided & 0.68$^{\dagger}$ & 0.73$^{\dagger}$ & 0.62$^{\dagger}$ & 0.58$^{\dagger}$ & 0.76 & 0.76 & 0.76 & 0.77 & 0.77 & 0.77 & 0.79 \\
\addlinespace[2pt]
RADAR & 0.87 & 1-sided & 0.86 & 0.87 & 0.91$^{*}$ & 0.96$^{*}$ & 0.86 & 0.88 & 0.90$^{*}$ & 0.86 & 0.87 & 0.87 & 0.88 \\
& & 2-sided & 0.79$^{\dagger}$ & 0.85$^{\dagger}$ & 0.57$^{\dagger}$ & 0.57$^{\dagger}$ & 0.87 & 0.87 & 0.84$^{\dagger}$ & 0.88 & 0.88 & 0.88 & 0.88 \\
\addlinespace[2pt]
RoBERTa & 0.72 & 1-sided & 0.81$^{*}$ & 0.70 & 0.08 & 0.02 & 0.07 & 0.67 & 0.39 & 0.69 & 0.72 & 0.71 & 0.72 \\
& & 2-sided & 0.65$^{\dagger}$ & 0.71 & 0.55$^{\dagger}$ & 0.47$^{\dagger}$ & 0.64$^{\dagger}$ & 0.74 & 0.68$^{\dagger}$ & 0.73 & 0.73 & 0.73 & 0.73 \\
\bottomrule
\end{tabular}
\caption{AUROC on RAID, every detector against every attack. \textbf{1-sided} scores attacked AI text against clean human text, which is what the standard protocol reports. \textbf{2-sided} scores it against attacked human text. $^{*}$ marks a significant increase over clean, $^{\dagger}$ a significant decrease. Every starred one-sided gain sits above a daggered two-sided loss, and no pair increases significantly two-sided.}
\label{tab:twosided}
\end{table*}

\begin{table*}[t]
\centering\small
\setlength{\tabcolsep}{3.4pt}
\begin{tabular}{@{}lcccccccccccc@{}}
\toprule
& \multicolumn{12}{c}{TPR@1\%FPR -- fraction of AI documents detected} \\
\cmidrule(l){2-13}
detector & none & paraphr. & synonym & homogl. & zero-w. & whitesp. & up/low & ins-par & art-del & alt-sp. & misspell & number \\
\midrule
Binoculars & 0.74 & 0.68 & 0.30 & 0.16 & 0.80 & 0.62 & 0.67 & 0.72 & 0.67 & 0.72 & 0.72 & 0.71 \\
FDG-neo & 0.69 & 0.59 & 0.22 & 0.08 & 0.06 & 0.49 & 0.52 & 0.67 & 0.57 & 0.67 & 0.65 & 0.61 \\
FDG-gpt2xl & 0.58 & 0.53 & 0.23 & 0.23 & 0.03 & 0.37 & 0.40 & 0.57 & 0.47 & 0.56 & 0.53 & 0.53 \\
FDG-gpt2 & 0.38 & 0.26 & 0.11 & 0.61 & 0.00 & 0.09 & 0.15 & 0.33 & 0.20 & 0.36 & 0.33 & 0.34 \\
Lastde & 0.41 & 0.18 & 0.09 & 0.69 & 0.22 & 0.37 & 0.16 & 0.36 & 0.27 & 0.36 & 0.34 & 0.31 \\
LogLik & 0.21 & 0.06 & 0.09 & 0.11 & 0.95 & 0.14 & 0.16 & 0.19 & 0.17 & 0.20 & 0.19 & 0.16 \\
LogRank & 0.23 & 0.07 & 0.09 & 0.10 & 0.79 & 0.16 & 0.16 & 0.21 & 0.18 & 0.22 & 0.22 & 0.19 \\
RADAR & 0.24 & 0.21 & 0.16 & 0.00 & 0.00 & 0.13 & 0.19 & 0.21 & 0.19 & 0.23 & 0.22 & 0.23 \\
RoBERTa & 0.23 & 0.19 & 0.17 & 0.00 & 0.00 & 0.00 & 0.11 & 0.01 & 0.16 & 0.21 & 0.21 & 0.22 \\
\bottomrule
\end{tabular}
\caption{TPR at a 1\% false-positive threshold, every detector $\times$ every attack. Cells are the fraction of \emph{attacked AI} documents scoring above the detector's threshold. Truncation cells are reported separately in Section~\ref{sec:length}.}
\label{tab:tprgrid}
\end{table*}

\begin{table*}[t]
\centering\small
\setlength{\tabcolsep}{3.4pt}
\begin{tabular}{@{}lcccccccccccc@{}}
\toprule
& \multicolumn{12}{c}{FPR -- fraction of \emph{human} documents wrongly flagged} \\
\cmidrule(l){2-13}
detector & none & paraphr. & synonym & homogl. & zero-w. & whitesp. & up/low & ins-par & art-del & alt-sp. & misspell & number \\
\midrule
Binoculars & 0.01 & 0.14 & 0.00 & 0.00 & 0.62 & 0.00 & 0.00 & 0.01 & 0.00 & 0.01 & 0.00 & 0.01 \\
FDG-neo & 0.01 & 0.20 & 0.00 & 0.01 & 0.02 & 0.00 & 0.00 & 0.01 & 0.00 & 0.01 & 0.00 & 0.01 \\
FDG-gpt2xl & 0.01 & 0.15 & 0.00 & 0.13 & 0.00 & 0.01 & 0.00 & 0.01 & 0.00 & 0.01 & 0.00 & 0.01 \\
FDG-gpt2 & 0.01 & 0.07 & 0.00 & 0.70 & 0.00 & 0.00 & 0.00 & 0.01 & 0.00 & 0.00 & 0.00 & 0.01 \\
Lastde & 0.01 & 0.01 & 0.00 & 0.71 & 0.10 & 0.01 & 0.00 & 0.01 & 0.01 & 0.02 & 0.01 & 0.00 \\
LogLik & 0.01 & 0.00 & 0.00 & 0.00 & 0.98 & 0.00 & 0.00 & 0.01 & 0.01 & 0.01 & 0.01 & 0.00 \\
LogRank & 0.01 & 0.00 & 0.00 & 0.00 & 0.81 & 0.00 & 0.00 & 0.01 & 0.01 & 0.01 & 0.01 & 0.00 \\
RADAR & 0.01 & 0.03 & 0.01 & 0.00 & 0.00 & 0.01 & 0.01 & 0.02 & 0.01 & 0.01 & 0.01 & 0.01 \\
RoBERTa & 0.01 & 0.07 & 0.01 & 0.00 & 0.00 & 0.00 & 0.00 & 0.00 & 0.00 & 0.01 & 0.01 & 0.01 \\
\bottomrule
\end{tabular}
\caption{False-positive rate on \emph{attacked human} text at the same 1\%-FPR thresholds. Cells are the fraction of \emph{attacked human} documents scoring above the same threshold. The \texttt{none} column is the nominal rate by construction.}
\label{tab:fprgrid}
\end{table*}

\begin{table*}[t]
\centering\small
\setlength{\tabcolsep}{5pt}
\begin{tabular}{@{}lllll@{}}
\toprule
name & scoring model(s) & size & family & source \\
\midrule
Binoculars & Falcon-7B + Falcon-7B-Instruct & 7B$\times$2 & modern zero-shot & \citet{hans2024binoculars} \\
FDG-neo & Fast-DetectGPT / GPT-Neo & 2.7B & modern zero-shot & \citet{bao2024fastdetectgpt} \\
FDG-gpt2xl & Fast-DetectGPT / GPT-2 XL & 1.5B & modern zero-shot & \citet{bao2024fastdetectgpt} \\
FDG-gpt2 & Fast-DetectGPT / GPT-2 & 124M & modern zero-shot & \citet{bao2024fastdetectgpt} \\
Lastde & Lastde / GPT-Neo & 2.7B & modern zero-shot & \citet{xu2025lastde} \\
LogLik & mean log-likelihood / GPT-Neo & 2.7B & legacy statistical & \citet{gehrmann2019gltr} \\
LogRank & mean log-rank / GPT-Neo & 2.7B & legacy statistical & \citet{gehrmann2019gltr} \\
RADAR & RADAR-Vicuna-7B & 355M & adversarially trained & \citet{hu2023radar} \\
RoBERTa & HC3 ChatGPT-detector & 125M & supervised classifier & \citet{guo2023hc3} \\
\bottomrule
\end{tabular}
\caption{The nine detectors. All run from public checkpoints without
modification; scores are normalized so that higher means more AI-like.}
\label{tab:detectors}
\end{table*}

\begin{table*}[t]
\centering\small
\begin{tabular}{@{}lllrlr@{}}
\toprule
detector & best generator & worst generator & gap & worst attack & loss \\
\midrule
Binoculars & ChatGPT (1.000) & MPT (0.430) & \textbf{0.570} & synonym & 0.181 \\
FDG-neo & GPT-4 (1.000) & MPT (0.425) & \textbf{0.575} & homoglyph & 0.304 \\
FDG-gpt2xl & GPT-2 (1.000) & Mistral (0.373) & \textbf{0.627} & zero-width & 0.321 \\
FDG-gpt2 & GPT-3 (0.973) & Mistral (0.368) & \textbf{0.605} & zero-width & 0.248 \\
Lastde & GPT-2 (0.999) & GPT-4 (0.400) & \textbf{0.599} & synonym & 0.290 \\
LogLik & GPT-2 (0.978) & MPT (0.334) & \textbf{0.644} & synonym & 0.238 \\
LogRank & GPT-2 (0.988) & MPT (0.348) & \textbf{0.640} & synonym & 0.236 \\
RADAR & GPT-4 (0.971) & Mistral (0.597) & \textbf{0.374} & whitespace & 0.017 \\
RoBERTa & Mistral-Chat (0.943) & MPT (0.321) & 0.622 & zero-width & \textbf{0.705} \\
\bottomrule
\end{tabular}
\caption{Generator choice against the worst attack, clean AUROC on RAID.
Best and worst are each detector's highest and lowest AUROC over the 22
generator and decoding setups, and gap is their difference. Every best
uses greedy decoding and every worst uses sampling. Loss is the largest
drop any attack causes. Bold marks the larger of gap and loss.}
\label{tab:generators}
\end{table*}

\begin{figure}[t]
  \centering
  \includegraphics[width=\columnwidth]{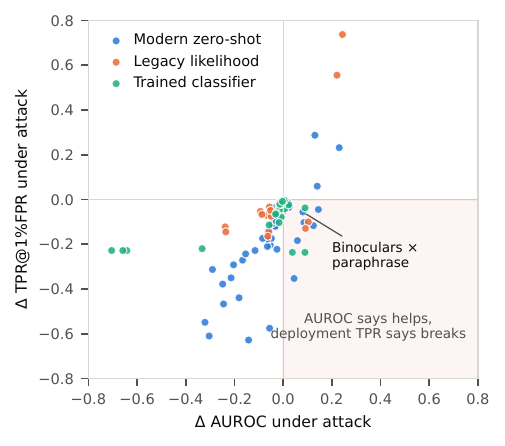}
  \caption{$\Delta$AUROC vs.\ $\Delta$TPR@1\%FPR for all 99 cells. Points in
  the shaded quadrant are cells where AUROC reports improvement while
  deployment-relevant TPR degrades; for Binoculars$\times$paraphrase both
  deltas are individually significant, with opposite signs.}
  \label{fig:divergence}
\end{figure}

\section{DetectRL result grids}
\label{sec:grids-dr}

Tables~\ref{tab:twosided-dr}, \ref{tab:tprgrid-dr} and \ref{tab:fprgrid-dr}
show the repeated experiment results with the three grids on DetectRL. Six of its seven attacks are its own, so the columns do not match the RAID grids. But it shows us the same pattern we found on RAID.

Table~\ref{tab:twosided-dr} has the same layout as Table~\ref{tab:twosided}.
Every significant one-sided gain sits above a significant two-sided loss, as
on RAID. No pair gains significantly two-sided.
Table~\ref{tab:fprgrid-dr} shows that under zero-width insertion, log-likelihood and log-rank flag \emph{every one} of the $1{,}440$ attacked human documents. RADAR flags $37\%$ of them under back-translation, which the standard protocol scores as a significant improvement.

\begin{table*}[t]
\centering\small
\setlength{\tabcolsep}{3.4pt}
\begin{tabular}{@{}lcccccccc@{}}
\toprule
& \multicolumn{8}{c}{TPR@1\%FPR -- fraction of AI documents detected} \\
\cmidrule(l){2-9}
detector & none & DIPPER & back-tr. & polish & word & char & homogl. & zero-w. \\
\midrule
Binoculars & 0.73 & 0.78 & 0.09 & 0.60 & 0.21 & 0.25 & 0.26 & 0.36 \\
FDG-neo & 0.65 & 0.72 & 0.07 & 0.48 & 0.08 & 0.15 & 0.11 & 0.06 \\
FDG-gpt2xl & 0.57 & 0.58 & 0.07 & 0.39 & 0.05 & 0.06 & 0.03 & 0.00 \\
FDG-gpt2 & 0.40 & 0.37 & 0.04 & 0.22 & 0.02 & 0.04 & 0.05 & 0.00 \\
Lastde & 0.56 & 0.63 & 0.04 & 0.31 & 0.00 & 0.00 & 0.01 & 0.70 \\
LogLik & 0.42 & 0.37 & 0.01 & 0.22 & 0.00 & 0.00 & 0.00 & 1.00 \\
LogRank & 0.46 & 0.45 & 0.01 & 0.25 & 0.00 & 0.00 & 0.00 & 1.00 \\
RADAR & 0.41 & 0.55 & 0.43 & 0.33 & 0.28 & 0.29 & 0.28 & 0.02 \\
RoBERTa & 0.17 & 0.17 & 0.07 & 0.11 & 0.02 & 0.00 & 0.00 & 0.00 \\
\bottomrule
\end{tabular}
\caption{TPR at a 1\% false-positive threshold on DetectRL. Six of the seven attacks are DetectRL's own; zero-width is RAID's, applied by us because DetectRL has no equivalent.}
\label{tab:tprgrid-dr}
\end{table*}

\begin{table*}[t]
\centering\small
\setlength{\tabcolsep}{3.4pt}
\begin{tabular}{@{}lcccccccc@{}}
\toprule
detector & none & DIPPER & back-tr. & polish & word & char & homogl. & zero-w. \\
\midrule
Binoculars & 0.01 & 0.08 & 0.00 & 0.14 & 0.00 & 0.00 & 0.00 & 0.31 \\
FDG-neo & 0.01 & 0.14 & 0.00 & 0.13 & 0.00 & 0.00 & 0.00 & 0.01 \\
FDG-gpt2xl & 0.01 & 0.07 & 0.00 & 0.12 & 0.00 & 0.00 & 0.00 & 0.00 \\
FDG-gpt2 & 0.01 & 0.04 & 0.00 & 0.07 & 0.00 & 0.01 & 0.00 & 0.00 \\
Lastde & 0.01 & 0.12 & 0.00 & 0.09 & 0.00 & 0.00 & 0.00 & 0.68 \\
LogLik & 0.01 & 0.02 & 0.00 & 0.06 & 0.00 & 0.00 & 0.00 & 1.00 \\
LogRank & 0.01 & 0.02 & 0.00 & 0.05 & 0.00 & 0.00 & 0.00 & 1.00 \\
RADAR & 0.01 & 0.12 & 0.37 & 0.09 & 0.01 & 0.00 & 0.00 & 0.02 \\
RoBERTa & 0.01 & 0.01 & 0.00 & 0.05 & 0.00 & 0.00 & 0.00 & 0.00 \\
\bottomrule
\end{tabular}
\caption{False-positive rate on \emph{attacked human} text in DetectRL, at the same 1\%-FPR thresholds. Log-likelihood and log-rank flag every one of the 1{,}440 human documents under zero-width insertion.}
\label{tab:fprgrid-dr}
\end{table*}

\begin{table*}[t]
\centering\footnotesize
\setlength{\tabcolsep}{2.0pt}
\begin{tabular}{@{}llcccccccc@{}}
\toprule
& \multicolumn{9}{c}{AUROC} \\
\cmidrule(l){2-10}
detector & clean & protocol & DIPPER & back-tr. & polish & word & char & homogl. & zero-w. \\
\midrule
Binoculars & 0.96 & 1-sided & 0.97$^{*}$ & 0.60 & 0.86 & 0.73 & 0.77 & 0.81 & 0.99$^{*}$ \\
& & 2-sided & 0.91$^{\dagger}$ & 0.76$^{\dagger}$ & 0.75$^{\dagger}$ & 0.86$^{\dagger}$ & 0.88$^{\dagger}$ & 0.89$^{\dagger}$ & 0.55$^{\dagger}$ \\
\addlinespace[2pt]
FDG-neo & 0.90 & 1-sided & 0.96$^{*}$ & 0.46 & 0.79 & 0.51 & 0.55 & 0.54 & 0.67 \\
& & 2-sided & 0.87$^{\dagger}$ & 0.76$^{\dagger}$ & 0.70$^{\dagger}$ & 0.83$^{\dagger}$ & 0.80$^{\dagger}$ & 0.83$^{\dagger}$ & 0.66$^{\dagger}$ \\
\addlinespace[2pt]
FDG-gpt2xl & 0.88 & 1-sided & 0.93$^{*}$ & 0.54 & 0.77 & 0.50 & 0.47 & 0.43 & 0.45 \\
& & 2-sided & 0.84$^{\dagger}$ & 0.81$^{\dagger}$ & 0.68$^{\dagger}$ & 0.84$^{\dagger}$ & 0.77$^{\dagger}$ & 0.80$^{\dagger}$ & 0.63$^{\dagger}$ \\
\addlinespace[2pt]
FDG-gpt2 & 0.83 & 1-sided & 0.86$^{*}$ & 0.45 & 0.69 & 0.42 & 0.44 & 0.47 & 0.41 \\
& & 2-sided & 0.76$^{\dagger}$ & 0.76$^{\dagger}$ & 0.63$^{\dagger}$ & 0.78$^{\dagger}$ & 0.64$^{\dagger}$ & 0.70$^{\dagger}$ & 0.39$^{\dagger}$ \\
\addlinespace[2pt]
Lastde & 0.90 & 1-sided & 0.94$^{*}$ & 0.41 & 0.77 & 0.20 & 0.29 & 0.33 & 0.97$^{*}$ \\
& & 2-sided & 0.84$^{\dagger}$ & 0.73$^{\dagger}$ & 0.67$^{\dagger}$ & 0.78$^{\dagger}$ & 0.77$^{\dagger}$ & 0.80$^{\dagger}$ & 0.57$^{\dagger}$ \\
\addlinespace[2pt]
LogLik & 0.91 & 1-sided & 0.92 & 0.25 & 0.81 & 0.42 & 0.32 & 0.42 & 1.00$^{*}$ \\
& & 2-sided & 0.86$^{\dagger}$ & 0.85$^{\dagger}$ & 0.68$^{\dagger}$ & 0.90 & 0.88$^{\dagger}$ & 0.89 & 0.71$^{\dagger}$ \\
\addlinespace[2pt]
LogRank & 0.91 & 1-sided & 0.92 & 0.27 & 0.80 & 0.46 & 0.33 & 0.45 & 1.00$^{*}$ \\
& & 2-sided & 0.86$^{\dagger}$ & 0.84$^{\dagger}$ & 0.69$^{\dagger}$ & 0.90 & 0.88$^{\dagger}$ & 0.89 & 0.70$^{\dagger}$ \\
\addlinespace[2pt]
RADAR & 0.92 & 1-sided & 0.95$^{*}$ & 0.96$^{*}$ & 0.90 & 0.89 & 0.93 & 0.92 & 0.97$^{*}$ \\
& & 2-sided & 0.83$^{\dagger}$ & 0.55$^{\dagger}$ & 0.77$^{\dagger}$ & 0.88$^{\dagger}$ & 0.91 & 0.91 & 0.48$^{\dagger}$ \\
\addlinespace[2pt]
RoBERTa & 0.84 & 1-sided & 0.86$^{*}$ & 0.79 & 0.77 & 0.69 & 0.64 & 0.63 & 0.08 \\
& & 2-sided & 0.77$^{\dagger}$ & 0.75$^{\dagger}$ & 0.58$^{\dagger}$ & 0.83 & 0.80 & 0.80 & 0.50$^{\dagger}$ \\
\bottomrule
\end{tabular}
\caption{AUROC on DetectRL, every detector against every attack.
\textbf{1-sided} scores attacked AI text against clean human text.
\textbf{2-sided} scores it against attacked human text. $^{*}$ marks a
significant increase over clean, $^{\dagger}$ a significant decrease. No cell increases significantly two-sided.}
\label{tab:twosided-dr}
\end{table*}

\begin{table}[t]
\centering\small
\begin{tabular}{@{}lcc@{}}
\toprule
attack & semantic sim. & positions changed \\
\midrule
synonym & 0.917 & 20.9\% \\
paraphr. & 0.874 & 96.1\% \\
art-del & 0.991 & 81.6\% \\
up/low & 1.000 & 4.7\% \\
ins-par & 1.000 & 1.5\% \\
number & 0.997 & 1.1\% \\
misspell & 0.994 & 1.0\% \\
alt-sp. & 0.994 & 0.7\% \\
\bottomrule
\end{tabular}
\caption{Semantic meaning preserved by word-level attacks determined by cosine similarity of the attacked text and its source, over $2{,}160$ documents per attack. Positions changed is the share of word positions that differ from the source. Unicode and whitespace attacks are omitted since their scores depend on the tokenizer, not on meaning (Table~\ref{tab:fidelity}).}
\label{tab:semantic}
\end{table}

\begin{figure*}[t]
  \centering
  \includegraphics[width=\textwidth]{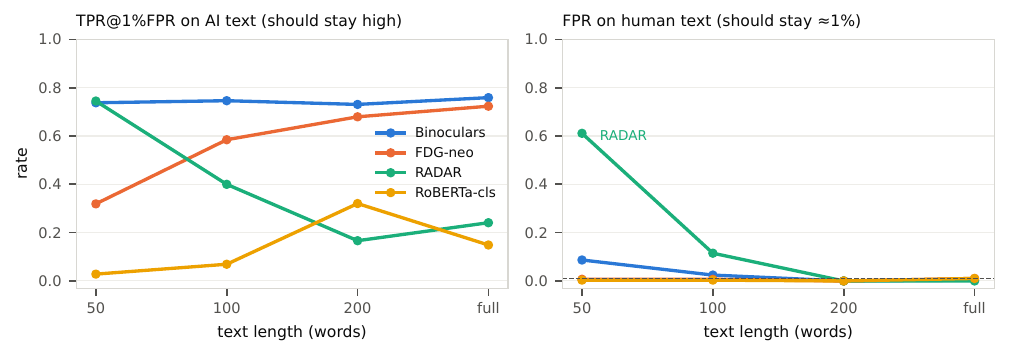}
  \caption{Controlled truncation of identical sources. Left: detection at
  1\%-FPR thresholds. Right: false-positive rate on human text at the same
  thresholds. RADAR's apparent short-text gain is length bias: it flags 61\%
  of 50-word human texts.}
  \label{fig:length}
\end{figure*}

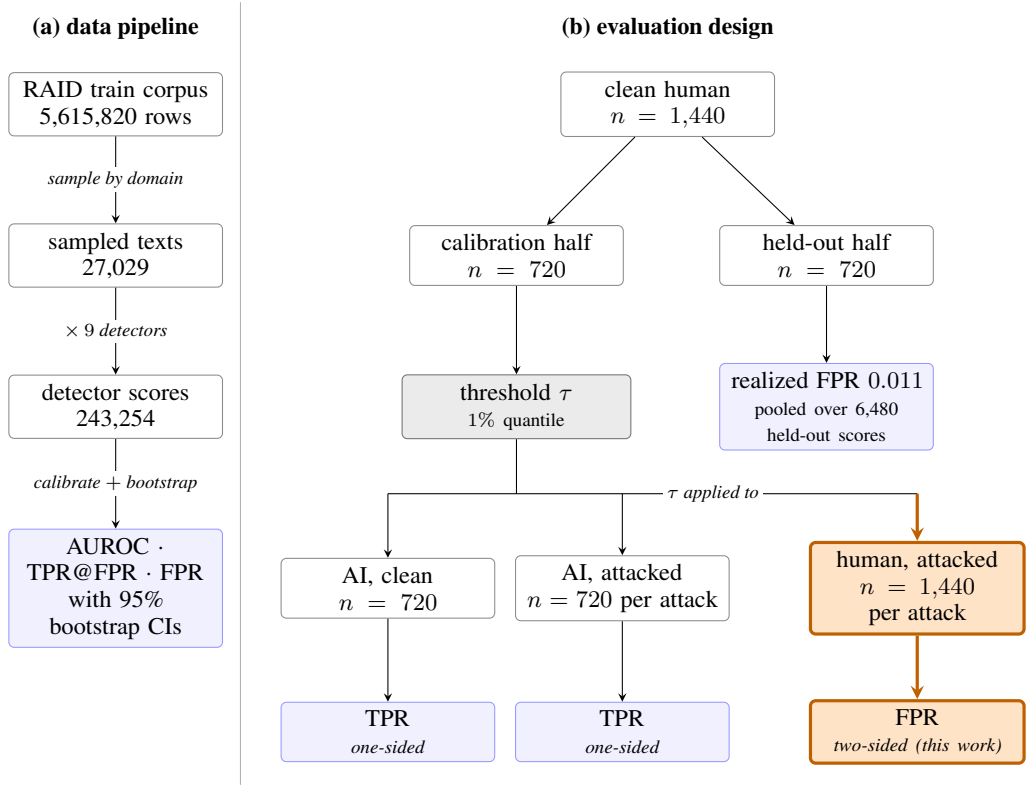
\begin{figure*}[t]
\centering
\begin{tikzpicture}[
  font=\footnotesize,
  node distance=0pt,
  bx/.style={draw, rounded corners=2pt, align=center, inner sep=3.2pt,
             minimum height=7mm, text width=#1},
  dat/.style={bx=26mm, draw=black!45},
  res/.style={bx=26mm, fill=blue!6, draw=blue!40},
  new/.style={bx=26mm, fill=orange!22, draw=orange!75!black, very thick},
  lbl/.style={font=\scriptsize\itshape, inner sep=1.5pt},
  ar/.style={->, >=stealth, thin},
]
\node[dat] (a1) at (0,0) {RAID train corpus\\5{,}615{,}820 rows};
\node[dat] (a2) at (0,-2.0) {sampled texts\\27{,}029};
\node[dat] (a3) at (0,-4.0) {detector scores\\243{,}254};
\node[res] (a4) at (0,-6.4) {AUROC $\cdot$ TPR@FPR $\cdot$ FPR\\with 95\% bootstrap CIs};
\draw[ar] (a1) -- node[lbl, fill=white] {sample by domain} (a2);
\draw[ar] (a2) -- node[lbl, fill=white] {$\times\,9$ detectors} (a3);
\draw[ar] (a3) -- node[lbl, fill=white] {calibrate $+$ bootstrap} (a4);
\node[font=\small\bfseries] at (0,1.0) {(a) data pipeline};
\node[dat] (h) at (7.3,0) {clean human\\$n=1{,}440$};
\node[dat] (hc) at (5.3,-2.0) {calibration half\\$n=720$};
\node[dat] (he) at (9.4,-2.0) {held-out half\\$n=720$};
\node[bx=28mm, fill=black!8, draw=black!55] (th) at (5.3,-4.0)
  {threshold $\tau$\\{\scriptsize $1\%$ quantile}};
\node[res] (rf) at (9.4,-4.0)
  {realized FPR $0.011$\\{\scriptsize pooled over
  6{,}480 held-out scores}};
\draw[ar] (h) -- (hc);  \draw[ar] (h) -- (he);
\draw[ar] (hc) -- (th); \draw[ar] (he) -- (rf);
\draw[thin] (th) -- (5.3,-5.15);
\draw[thin] (3.6,-5.15) -- (10.6,-5.15);
\node[lbl, fill=white] at (7.9,-5.15) {$\tau$ applied to};
\node[dat] (s1) at (3.6,-6.4) {AI, clean\\$n=720$};
\node[dat] (s2) at (6.7,-6.4) {AI, attacked\\$n=720$ per attack};
\node[new] (s3) at (10.6,-6.4) {human, attacked\\$n=1{,}440$ per attack};
\draw[ar] (3.6,-5.15) -- (s1);
\draw[ar] (6.7,-5.15) -- (s2);
\draw[ar, orange!75!black, very thick] (10.6,-5.15) -- (s3);
\node[res] (r1) at (3.6,-8.3) {TPR\\{\scriptsize\itshape one-sided}};
\node[res] (r2) at (6.7,-8.3) {TPR\\{\scriptsize\itshape one-sided}};
\node[new] (r3) at (10.6,-8.3) {FPR\\{\scriptsize\itshape two-sided (this work)}};
\draw[ar] (s1) -- (r1); \draw[ar] (s2) -- (r2);
\draw[ar, orange!75!black, very thick] (s3) -- (r3);
\node[font=\small\bfseries] at (7.3,1.0) {(b) evaluation design};
\draw[black!30] (1.65,1.35) -- (1.65,-9.0);
\end{tikzpicture}
\caption{The measurement pipeline. \textbf{(a)} We sample RAID by domain and score every text with all 9 detectors (Similar approach for DetectRL). \textbf{(b)} Clean human documents are split in half. One half fits each detector's threshold $\tau$ at a nominal
$1\%$ false-positive rate, the other verifies it works. Attacked text never enters calibration, so the attacked boxes carry the full $n$. Standard evaluations score only the two AI branches. The \colorbox{orange!22}{highlighted branch} is what this study adds, applying the same 11 attacks to \emph{human} documents and counting who gets accused.}
\label{fig:pipeline}
\end{figure*}

\section{DetectRL sampling and a labeling defect}
\label{sec:detectrl-detail}

We sample DetectRL's source corpus rather than its prepared splits. Each source record contains a human document, a machine document, and six attacked variants of each, so every attack is paired with its clean original. We draw 360 human and 180 machine sources per domain, seeded, retaining only complete records with texts of at least 30 words. This matches RAID sample sizes.

Prepared splits show a labeling defect. In one evaluation set, 336 of $1{,}008$ records labelled as human actually contain machine text. The corresponding training set shows the same issue, with $10{,}847$ of $32{,}574$ records mislabeled.

We confirmed this by matching affected records back to the source corpus and by measuring character trigram overlap. Fields designated as human align strongly with human text, while those designated as machine align strongly with machine text, indicating that mislabelling occurred during split construction. The problem is limited to those sets. Our sampler repeats the trigram check each run and rejects samples if it fails.

\section{The curated confusables table}
\label{sec:curated-table}

The curated table holds 47 mappings onto 34 ASCII targets, drawn from Cyrillic (26 sources), Greek (19) and two others. It covers the characters
RAID's homoglyph attacker substitutes, which are almost entirely Cyrillic and Greek lookalikes of Latin letters. Unicode's full UTS~\#39 list has 1{,}861 ASCII-target mappings by comparison. Table~\ref{tab:confusables-tradeoff} gives both tables' coverage on each corpus, and
Table~\ref{tab:sanitizer-detectors} the per-detector recovery that Table~\ref{tab:variants} reports as ranges.

\begin{table}[t]
\centering\small
\setlength{\tabcolsep}{4pt}
\begin{tabular}{@{}llrrr@{}}
\toprule
corpus & table & mappings & reversed & clean rewritten \\
\midrule
RAID & curated & 47 & 99.3\% & 60 (2.8\%) \\
RAID & full & 1,861 & 99.9\% & 330 (15.3\%) \\
DetectRL & curated & 47 & 61.8\% & 17 (0.8\%) \\
DetectRL & full & 1,861 & 98.0\% & 44 (2.0\%) \\
\bottomrule
\end{tabular}
\caption{The curated table against the full UTS~\#39 list. Reversed is the percentage of the attack's substituted characters turned back into the original letter. Clean rewritten is how many of the $2{,}160$ unattacked source documents the table alters.}
\label{tab:confusables-tradeoff}
\end{table}

\begin{table*}[t]
\centering\small
\setlength{\tabcolsep}{4pt}
\begin{tabular}{@{}lrrrrrrrrr@{}}
\toprule
detector & \multicolumn{3}{c}{zero-width} & \multicolumn{3}{c}{homoglyph} & \multicolumn{3}{c}{whitespace} \\
\cmidrule(lr){2-4}\cmidrule(lr){5-7}\cmidrule(l){8-10}
 & curated & full & cur.+ws & curated & full & cur.+ws & curated & full & cur.+ws \\
\midrule
Binoculars & 100 & 100$^{\dagger}$ & 100 & 97 & 98$^{\dagger}$ & 97 & 84 & 100$^{\dagger}$ & 99 \\
FDG-neo & 100 & 100 & 100 & 95 & 95 & 95 & 70 & 99 & 99 \\
FDG-gpt2xl & 100 & 100$^{\dagger}$ & 99 & 92 & 92$^{\dagger}$ & 91 & 64 & 99$^{\dagger}$ & 99 \\
FDG-gpt2 & 100 & 100$^{\dagger}$ & 100 & 90 & 93$^{\dagger}$ & 90 & 23 & 99$^{\dagger}$ & 99 \\
Lastde & 100 & 100 & 100 & 87 & 80 & 88 & 92 & 99 & 99 \\
LogLik & 100 & 100 & 100 & 93 & 86 & 92 & 67 & 100 & 100 \\
LogRank & 100 & 100 & 100 & 86 & 82 & 86 & 67 & 100 & 100 \\
RADAR & 100 & 100 & 99 & 104 & 105 & 103 & 56 & 101 & 100 \\
RoBERTa & 100 & 102$^{\dagger}$ & 100$^{\dagger}$ & 94 & 81$^{\dagger}$ & 94$^{\dagger}$ & 0 & 92$^{\dagger}$ & 90$^{\dagger}$ \\
\bottomrule
\end{tabular}
\caption{Detection after each sanitizer configuration, per detector, as a percentage of clean detection at the $1\%$ threshold. Table~\ref{tab:variants} gives the range of each column. $^{\dagger}$ marks a configuration that raises the detector's FPR on \emph{clean} human text.}
\label{tab:sanitizer-detectors}
\end{table*}

\begin{table*}[t]
\centering\small
\setlength{\tabcolsep}{3pt}
\begin{tabular}{@{}llllll@{}}
\toprule
\multicolumn{6}{c}{target $\leftarrow$ source code points} \\
\midrule
\texttt{A} & U+0391, U+0410 & \texttt{S} & U+0405 & \texttt{k} & U+03BA \\
\texttt{B} & U+0392, U+0412 & \texttt{T} & U+03A4, U+0422 & \texttt{l} & U+217C \\
\texttt{C} & U+0421 & \texttt{X} & U+03A7, U+0425 & \texttt{o} & U+03BF, U+043E \\
\texttt{E} & U+0395, U+0415 & \texttt{Y} & U+03A5 & \texttt{p} & U+03C1, U+0440 \\
\texttt{H} & U+0397, U+041D & \texttt{Z} & U+0396 & \texttt{q} & U+051B \\
\texttt{I} & U+0399 & \texttt{a} & U+03B1, U+0430 & \texttt{s} & U+0455 \\
\texttt{J} & U+0408 & \texttt{c} & U+0441 & \texttt{v} & U+03BD \\
\texttt{K} & U+039A, U+041A & \texttt{d} & U+0501 & \texttt{w} & U+0461 \\
\texttt{M} & U+039C, U+041C & \texttt{e} & U+0435 & \texttt{x} & U+0445 \\
\texttt{N} & U+039D & \texttt{g} & U+0261 & \texttt{y} & U+0443 \\
\texttt{O} & U+039F, U+041E & \texttt{i} & U+0456 & & \\
\texttt{P} & U+03A1, U+0420 & \texttt{j} & U+0458 & & \\
\bottomrule
\end{tabular}
\caption{The 47 curated confusable mappings. Each row gives an ASCII target and the code points folded onto it. Applied after zero-width stripping and
before NFKC.}
\label{tab:curated}

\end{table*}


One gap is worth recording. RAID replaces capital \texttt{I} with either Greek \textsc{u+0399} or Cyrillic \textsc{u+0406}, and our table covers only
the first. Everything else folds back correctly. So a document with a capital \texttt{I} that drew the Cyrillic variant comes back readable but not byte-for-byte identical. That is why homoglyph documents restore exactly
$31\%$ of the time in Table~\ref{tab:fidelity} while their trigram similarity reaches $0.988$.

\section{How the $1\%$ threshold is set}
\label{sec:calibration}
The clean human documents are split in half, a calibration half $H_{\mathrm{cal}}$ and a held-out half $H_{\mathrm{eval}}$, each of 720. The threshold is the 99th percentile of the calibration scores,
\begin{equation}
\tau = Q_{0.99}\!\left(S(H_{\mathrm{cal}})\right),
\end{equation}

and every rate in the paper is the share of a set that scores above it,

\begin{align}
\mathrm{FPR} &= \tfrac{1}{|H_{\mathrm{eval}}|}\textstyle\sum_{x \in H_{\mathrm{eval}}} \mathbf{1}[S(x) > \tau], \\
\mathrm{TPR} &= \tfrac{1}{|M|}\textstyle\sum_{x \in M} \mathbf{1}[S(x) > \tau], \\
\mathrm{FPR}_a &= \tfrac{1}{|H_a|}\textstyle\sum_{x \in H_a} \mathbf{1}[S(x) > \tau].
\end{align}

For Binoculars on RAID, $\tau = -0.869$. It flags 6 of the 720 held-out clean human documents, $\mathrm{FPR} = 0.008$ against the $0.010$ asked
for. It flags 531 of the 720 clean machine documents, $\mathrm{TPR} = 0.74$, the entry in Table~\ref{tab:paraphrase}. Under zero-width insertion it flags 886 of the $1{,}440$ attacked human documents, $\mathrm{FPR}_a = 0.62$, the Binoculars entry in Figure~\ref{fig:fpr}. The same $\tau$ produced all three. Only the text changed. The bootstrap recomputes $\tau$ inside each of its $1{,}000$ replicates.

\section{Reproducibility}
We will release all scores, sampling manifests, seeds and analysis code. The scores total $724{,}096$, made up of $270{,}289$ RAID, $172{,}800$ on DetectRL and $281{,}007$ re-scores under three sanitizer settings. Bootstrap
intervals use $1{,}000$ replicates at seed 42. Compute was about $3$ GPU-hours on one H100 for the RAID sweep, $2$ for DetectRL, $3$ for the sanitized re-scores, and $1.5$ for the Falcon-7b backbone runs.

\end{document}